\documentclass[runningheads]{llncs}
\usepackage[T1]{fontenc}
\usepackage{graphicx}
\usepackage{amsmath,amssymb}
\usepackage{booktabs}
\usepackage{xcolor}
\usepackage{multirow}
\usepackage{color}
\usepackage{url}

\graphicspath{{../figures/}}
\begin{document}
\title{When Chain-of-Thought Backfires:\\ Evaluating Prompt Sensitivity in Medical Language Models}
\titlerunning{Prompt Sensitivity in Medical Language Models}
\author{Binesh Sadanandan\inst{1} \and
Vahid Behzadan\inst{1}}
\authorrunning{B. Sadanandan and V. Behzadan}
\institute{SAIL Lab, University of New Haven, West Haven, CT 06516, USA\\
\email{bsada1@unh.newhaven.edu, vbehzadan@newhaven.edu}}
\maketitle
\begin{abstract}
Large Language Models (LLMs) are increasingly deployed in medical settings, yet their sensitivity to prompt formatting remains poorly characterized. We evaluate MedGemma (4B and 27B parameters) on MedMCQA (4,183 questions) and PubMedQA (1,000 questions) across a broad suite of robustness tests. Our experiments reveal several concerning findings. Chain-of-Thought (CoT) prompting \textit{decreases} accuracy by 5.7\% compared to direct answering. Few-shot examples degrade performance by 11.9\% while increasing position bias from 0.14 to 0.47. Shuffling answer options causes the model to change predictions 59.1\% of the time, with accuracy dropping up to 27.4 percentage points. Front-truncating context to 50\% causes accuracy to plummet below the no-context baseline, yet back-truncation preserves 97\% of full-context accuracy. We further show that cloze scoring (selecting the highest log-probability option token) achieves 51.8\% (4B) and 64.5\% (27B), surpassing all prompting strategies and revealing that models ``know'' more than their generated text shows. Permutation voting recovers 4 percentage points over single-ordering inference. These results demonstrate that prompt engineering techniques validated on general-purpose models do not transfer to domain-specific medical LLMs, and that reliable alternatives exist.

\end{abstract}

\section{Introduction}

LLMs have achieved impressive performance on medical licensing exams, with GPT-4 exceeding passing thresholds by over 20 points~\cite{nori2023capabilities} and Med-PaLM~2 reaching 86.5\% on MedQA~\cite{singhal2023medpalm2}. These results have fueled enthusiasm for deploying LLMs in clinical decision support. But benchmark accuracy tells only part of the story. How models respond to variations in prompt format, question ordering, and context presentation remains poorly understood, despite being critical for real-world deployment where inputs are rarely formatted identically to benchmark conditions.

We focus on MedGemma~\cite{medgemma2025}, Google's medical-specialist LLM built on the Gemma architecture. A widely held belief in the LLM community is that certain prompting strategies reliably improve performance. CoT prompting, which instructs models to reason step-by-step before answering, has shown consistent gains on mathematical and logical reasoning tasks~\cite{wei2022chain}. Few-shot learning, where examples are provided in-context, helps models understand desired output formats~\cite{brown2020language}. These techniques are often treated as ``best practices'' that should transfer across domains and model families.

But should they? Domain-specific models may have internalized different patterns during training. A model trained extensively on medical literature might already encode structured clinical reasoning, making explicit CoT prompts redundant or even counterproductive. Similarly, few-shot examples drawn from one medical specialty might prime the model with concepts that are irrelevant or misleading for questions in other specialties.

We present a systematic evaluation of MedGemma's sensitivity to prompt variations across four experimental axes. First, we conduct a prompt ablation study comparing ten prompting strategies on 4,183 MedMCQA questions, measuring both accuracy and position bias. Second, we test option order sensitivity by shuffling answer choices across multiple random seeds and measuring how often the model changes its prediction. Third, we evaluate evidence conditioning on 1,000 PubMedQA questions across eleven context conditions, including five truncation strategies that reveal which parts of the abstract carry diagnostic value. Fourth, we benchmark robustness-oriented alternatives: cloze scoring, permutation voting, and CoT self-consistency. Our findings challenge conventional assumptions about prompt engineering in medical AI and identify practical mitigations for deployment.

\section{Related Work}

\paragraph{Medical Language Models and Benchmarks.}
Medical language models often demonstrate impressive benchmark scores on exam-style question answering. GPT-4 performs strongly on medical challenge sets~\cite{nori2023capabilities}, and Med-PaLM and Med-PaLM~2 report high scores on MedQA~\cite{singhal2023large,singhal2023medpalm2}. Open, domain-tuned models have also emerged, including MedGemma~\cite{medgemma2025} and BioMistral~\cite{labrak2024biomistral}. However, most reporting still emphasizes a single headline accuracy, while deployment inputs vary in formatting, context quality, and answer presentation.

\paragraph{Prompting for Reasoning.}
CoT prompting can improve performance on general reasoning tasks by eliciting intermediate steps~\cite{wei2022chain}. Follow-up work such as self-consistency explores sampling multiple reasoning paths and aggregating answers~\cite{wang2023selfconsistency}. These techniques are now common defaults, despite their added token cost and their sensitivity to output parsing.

\paragraph{When CoT Backfires.}
Recent work challenges the idea that CoT helps everywhere. Sprague et al.~\cite{sprague2024mind} identify settings where step-by-step prompting reduces accuracy, connecting these failures to cases where deliberate reasoning hurts humans as well. Meincke et al.\ argue that the gains from CoT have diminished for newer models and can even reverse on some tasks~\cite{meincke2024cot}. In medical question answering, Omar et al.\ compare multiple CoT-style prompts and find that improvements depend on the specific method and dataset, rather than following a simple monotonic trend~\cite{omar2024cotmedical}.

\paragraph{Prompt Sensitivity and Few-shot Formatting.}
Even without CoT, small prompt changes can shift performance. Lu et al.~\cite{lu2022fantastically} show that few-shot example order can materially affect accuracy, and Zhao et al.~\cite{zhao2021calibrate} propose calibration methods that reduce sensitivity to label and prompt priors. ProSA provides a more systematic view by measuring how model outputs vary across prompt templates~\cite{prosa2024}. Together, this work suggests that comparisons between models can be misleading if prompt choices are not controlled.

\paragraph{Multiple-choice Artifacts.}
Multiple-choice evaluation introduces its own failure modes. Zheng et al.~\cite{zheng2024llm} show that LLMs exhibit selection bias, preferring certain option identifiers even when content is balanced. Our option reordering experiments build on this line of work by separating changes in answer position from changes in distractor content.

\paragraph{Retrieval and Context Quality.}
Retrieval-Augmented Generation (RAG) augments a model with external documents at inference time. In medicine, benchmarking work finds large variation in RAG performance across retrievers and corpora, with a pronounced ``lost-in-the-middle'' effect in biomedical settings~\cite{xiong2024benchmarking,liu2024lost}. Retrieval can also introduce new failure modes: ClashEval shows that models can be led astray by incorrect retrieved context~\cite{wu2024clasheval}, and recent medical RAG methods aim to improve reliability under imperfect retrieval~\cite{sohn2024rag2,barnett2024rag}.

\section{Methods}

\subsection{Models and Datasets}

We evaluate MedGemma-4B, the 4-billion parameter instruction-tuned variant at bfloat16 precision, and MedGemma-27B, the 27-billion parameter model at full bfloat16 precision on 80\,GB A100 GPUs.

We use two standard medical QA benchmarks. MedMCQA~\cite{pal2022medmcqa} contains questions from Indian medical entrance examinations across 21 subjects; we use the 4,183-question validation split. PubMedQA~\cite{jin2019pubmedqa} contains research questions derived from PubMed titles that must be answered using abstracts; we use the 1,000-question labeled subset.

\subsection{Inference Protocol}

All experiments use a frozen inference protocol for reproducibility. Table~\ref{tab:inference} summarizes the two configurations.

\begin{table}[t]
\centering
\caption{Frozen inference parameters for all experiments.}
\label{tab:inference}
\begin{tabular}{lccp{3.8cm}}
\toprule
\textbf{Parameter} & \textbf{Determ.} & \textbf{Self-Cons.} & \textbf{Rationale} \\
\midrule
Temperature & 0.0 & 0.7 & Greedy for reproducibility; 0.7 for diverse reasoning paths \\
Top-$p$ & 1.0 & 0.95 & No nucleus filtering for greedy; mild filtering for sampling \\
Top-$k$ & 0 & 50 & Disabled for greedy; limits sampling to top 50 tokens \\
\texttt{do\_sample} & False & True & Greedy decoding vs.\ stochastic sampling \\
\texttt{max\_new\_tokens} & 256 & 256 & Sufficient for CoT reasoning while bounding cost \\
\bottomrule
\end{tabular}
\end{table}

All runs use the default system prompt and constrain outputs to the valid answer set (A through D for MedMCQA; yes/no/maybe for PubMedQA). We extract answers via regex; unparseable outputs are scored as incorrect. Where randomized perturbations are involved, we report results over multiple seeds (42, 123, 456) with mean and 95\% confidence intervals.

\subsection{Experimental Overview}

Table~\ref{tab:experiment_overview} summarizes the four experiments, their datasets, condition counts, and the core question each one addresses. Detailed descriptions follow below.

\begin{table}[t]
\centering
\caption{Overview of the four experiments in our evaluation.}
\label{tab:experiment_overview}
\begin{tabular}{clp{2.2cm}cp{3.4cm}}
\toprule
\textbf{Exp.} & \textbf{Name} & \textbf{Dataset} & \textbf{Conds.} & \textbf{Core Question} \\
\midrule
1 & Prompt Ablation & MedMCQA (4,183) & 10 & Do CoT and few-shot help or hurt? \\
2 & Option Order & MedMCQA (4,183) & 5 $\times$ 3 seeds & Does shuffling options change the answer? \\
3 & Evidence Cond. & PubMedQA (1,000) & 11 & How does context truncation affect accuracy? \\
4 & Robustness Basel. & MedMCQA (4,183) & 3 methods & Can alternative scoring mitigate fragility? \\
\bottomrule
\end{tabular}
\end{table}

\subsection{Experimental Conditions}

\paragraph{Experiment 1: Prompt Ablation.}
We test ten prompting strategies on MedMCQA. Five \textit{core} conditions vary the prompting approach: (1)~zero-shot direct; (2)~zero-shot CoT; (3)~few-shot direct with three curated examples; (4)~few-shot CoT with three curated examples; and (5)~answer-only (minimal prompt). Five additional conditions probe few-shot sensitivity: (6)~random selection from the training set; (7)~label-balanced selection (equal A/B/C/D representation); (8)~subject-matched selection (same medical specialty as the test item); and (9--10)~two alternative orderings of the curated examples to test order effects.

\paragraph{Experiment 2: Option Order Sensitivity.}
We apply four transformations to each MedMCQA question: random shuffle, rotate-1 (cyclic shift by one), rotate-2 (shift by two), and distractor swap (exchange incorrect options while preserving correct answer position). We measure the flip rate, defined as the proportion of questions where the model changes its answer when options are reordered. To quantify variance, we repeat the experiment across three random seeds and report mean accuracy and flip rate with 95\% confidence intervals.

\paragraph{Experiment 3: Evidence Conditioning.}
On PubMedQA, we vary context presentation across eleven conditions. Six \textit{core} conditions test context completeness: question-only (no context), full abstract, front-truncated 50\%, front-truncated 25\%, background-only (introduction sentences), and results-only (conclusion sentences). Five additional conditions test truncation strategy: back-truncated 50\% (keeping the final half), middle-truncated 50\% (keeping the central portion), sentence-boundary truncation at 50\%, and salient-sentence extraction retaining the top-5 and top-3 sentences ranked by non-stopword overlap with the question.

\paragraph{Experiment 4: Robustness Baselines.}
We evaluate three complementary scoring and aggregation methods on MedMCQA. \textit{Cloze scoring} bypasses free-text generation entirely: for each question we compute log-probabilities of the four option tokens (A through D) at the next-token position and select the highest. \textit{Permutation voting} runs greedy inference on multiple option orderings and takes a majority vote over inverse-mapped predictions. \textit{CoT self-consistency} samples $N$ CoT responses at temperature 0.7 and takes a majority vote, testing whether sampling diversity can overcome single-sample fragility.

\subsection{Metrics}

We report accuracy with 95\% Wilson score confidence intervals. Position bias is computed as a prediction frequency bias: $\text{Bias} = \frac{1}{4} \sum_{i \in \{A,B,C,D\}} |P_{\text{pred}}(i) - P_{\text{truth}}(i)|$, where $P_{\text{pred}}(i)$ is the fraction of model predictions for position $i$ and $P_{\text{truth}}(i)$ is the fraction of correct answers at position $i$. A score of 0 indicates the model's answer distribution perfectly matches the ground truth; higher values indicate systematic over- or under-selection of certain positions. For option-order experiments, we compute the flip rate: the proportion of questions where the model's prediction changes under reordering. Table~\ref{tab:metrics_summary} describes our additional metrics.

\begin{table}[t]
\centering
\caption{Additional metrics used in our evaluation.}
\label{tab:metrics_summary}
\begin{tabular}{lp{7.2cm}}
\toprule
\textbf{Metric} & \textbf{Definition} \\
\midrule
CKLD & $\sum_i (O_i - E_i)^2 / E_i$, where $O_i$ and $E_i$ are observed and expected counts per position; quantifies departure from uniform answering \\
RStd & $\sigma / \mu$ of per-position accuracies; captures how unevenly performance is distributed across answer positions \\
Per-position acc. & Accuracy conditioned on the correct answer being A, B, C, or D; reveals position-dependent competence \\
McNemar's test & Paired significance testing between conditions, with Bonferroni correction for multiple comparisons \\
\bottomrule
\end{tabular}
\end{table}

\section{Results}

\subsection{Prompt Ablation}

Table~\ref{tab:prompt_ablation} shows accuracy across prompting strategies for the five core conditions. Zero-shot direct achieves the highest accuracy at 47.6\%, while CoT \textit{reduces} accuracy by 5.7 percentage points (McNemar's test, $p < 0.001$). Few-shot examples cause an even larger degradation of 11.9\% ($p < 0.001$), while simultaneously increasing position bias from 0.137 to 0.472 (measured as prediction frequency bias, i.e., the mean absolute deviation between predicted and ground-truth answer distributions; see Section~3.5). All pairwise differences between zero-shot direct and other conditions remain statistically significant after Bonferroni correction ($p < 0.001$, McNemar's test). The model picks up spurious patterns from in-context examples rather than useful output formats.

\begin{table}[t]
\centering
\caption{Prompt ablation results on MedMCQA ($n{=}4{,}183$). Random baseline is 25\%.}
\label{tab:prompt_ablation}
\begin{tabular}{lccc}
\toprule
\textbf{Condition} & \textbf{Accuracy} & \textbf{95\% CI} & \textbf{Pos.\ Bias} \\
\midrule
Zero-shot direct & 47.6\% & [46.1, 49.1] & 0.137 \\
Zero-shot CoT & 41.9\% & [40.4, 43.3] & 0.275 \\
Few-shot direct (curated) & 35.7\% & [34.3, 37.0] & 0.472 \\
Few-shot CoT (curated) & 40.8\% & [39.4, 42.3] & 0.413 \\
Answer-only & 43.0\% & [41.5, 44.6] & 0.096 \\
\bottomrule
\end{tabular}
\end{table}

\subsection{Qualitative Examples}

Table~\ref{tab:sample_responses} shows representative model outputs that illustrate our key findings. The first example shows CoT backfiring: direct prompting answers ``D'' (Candidiasis) correctly, but CoT prompting starts analyzing each option, notes that tuberculosis is ``rare in children,'' and yet ultimately selects A (Tuberculosis). The second example shows an option-order flip: the model answers ``B'' (Anterior ethmoidal artery) under the original ordering, correctly identifying a non-branch of the external carotid. When options are shuffled, the content previously at position B moves to position A, but the model again answers ``B,'' now pointing to a different option entirely. It follows position, not content.

\begin{table}[t]
\centering
\caption{Sample model responses illustrating key failure modes. All examples from MedGemma-4B on MedMCQA. Prompts and outputs are reproduced verbatim.}
\label{tab:sample_responses}
\small
\begin{tabular}{p{1.8cm}p{8.7cm}}
\toprule
\multicolumn{2}{l}{\textbf{Example 1: CoT backfires} (Dental, correct answer: D)} \\
\midrule
Prompt (direct) & \texttt{Question: In a 6-month-old child, thick curd like white patch appears on the buccal mucosa. On rubbing it leaves an erythematous patch. Most likely diagnosis is:} \newline \texttt{Options: A) Tuberculosis\; B) Lichen planus\; C) Lupus erythematous\; D) Candidiasis} \newline \texttt{Answer with a single letter (A, B, C, or D).} \\
\addlinespace
Direct output & \texttt{D} \quad $\checkmark$ \\
\midrule
Prompt (CoT) & \texttt{Question: [same as above]} \newline \texttt{Think through this step-by-step, analyzing each option. Then provide your final answer as a single letter (A, B, C, or D).} \\
\addlinespace
CoT output & \texttt{A) Tuberculosis: Oral tuberculosis is rare in children. While it can present with lesions, they are less likely to be the classic "curd-like" appearance.} \newline \texttt{B) Lichen planus: less common in infants. It typically presents with flat-topped}\ldots \textit{(truncated at 256 tokens; selects} \texttt{A}\textit{)} \quad $\times$ \\
\midrule\midrule
\multicolumn{2}{l}{\textbf{Example 2: Option order flip} (Anatomy, correct answer: B $\to$ A after shuffle)} \\
\midrule
Prompt (orig.) & \texttt{Question: Which of the following are not a branch of external carotid artery in Kiesselbach's plexus.} \newline \texttt{Options: A) Sphenopalatine artery\; B) Anterior ethmoidal artery\; C) Greater palatine artery\; D) Septal branch of superior labial artery} \newline \texttt{Answer with a single letter (A, B, C, or D).} \\
\addlinespace
Output (orig.) & \texttt{B} \quad $\checkmark$ \quad (Anterior ethmoidal is correct; it is not an external carotid branch) \\
\midrule
Prompt (shuf.) & \texttt{Question: [same as above]} \newline \texttt{Options: A) Anterior ethmoidal artery\; B) Septal branch of superior labial artery\; C) Sphenopalatine artery\; D) Greater palatine artery} \\
\addlinespace
Output (shuf.) & \texttt{B} \quad $\times$ \quad (The correct content moved to position A, but the model still picks B) \\
\bottomrule
\end{tabular}
\end{table}

\subsection{Option Order Sensitivity}

Table~\ref{tab:option_order} reveals extreme sensitivity to option ordering. The mean flip rate is 59.1\%; the model changes its answer more often than not when options are shuffled. Rotation perturbations cause the largest accuracy drops (up to 27.4\%), while distractor swaps show smaller impact ($-$8.9\%). This confirms that position rather than distractor content drives the fragility.

\begin{table}[t]
\centering
\caption{Option order sensitivity on MedMCQA ($n{=}4{,}183$). All intervals are 95\% Wilson score CIs. Seed 42.}
\label{tab:option_order}
\begin{tabular}{lcccc}
\toprule
\textbf{Perturbation} & \textbf{Accuracy} & \textbf{95\% CI} & \textbf{Drop} & \textbf{Flip Rate} \\
\midrule
Original & 47.6\% & [46.1, 49.1] & -- & -- \\
Random shuffle & 27.3\% & [26.0, 28.7] & $-$20.3\% & 57.8\% \\
Rotate-1 & 20.2\% & [19.0, 21.5] & $-$27.4\% & 72.9\% \\
Rotate-2 & 21.9\% & [20.7, 23.2] & $-$25.7\% & 69.7\% \\
Distractor swap & 47.6\% & [46.1, 49.1] & $\phantom{-}$0.0\% & 36.1\% \\
\midrule
\textbf{Mean} & 29.2\% & -- & $-$18.4\% & 59.1\% \\
\bottomrule
\end{tabular}
\end{table}

\subsection{Evidence Conditioning}

Table~\ref{tab:evidence} shows that context presentation substantially affects PubMedQA performance. Most critically, front-truncated context performs \textit{worse} than no context: front-truncation at 50\% yields 13.8\% (4B) and 23.4\% (27B), far below question-only baselines of 34.5\% and 31.0\%. Naively truncating from the front actively misleads the model.

Our expanded truncation analysis reveals that \textit{how} context is truncated matters as much as \textit{how much} is removed. Back-truncation at 50\% preserves 97\% of full-context accuracy (44.5\% vs.\ 45.8\%), while front-truncation at the same ratio destroys it (13.8\%). The model relies on opening context for orientation; removing the beginning causes catastrophic failure, while removing the end is largely harmless.

Salient-sentence extraction offers a practical middle ground: selecting the top-5 sentences by question overlap achieves 40.1\%, retaining 88\% of full-context accuracy with roughly half the tokens.

\begin{table}[t]
\centering
\caption{Evidence conditioning on PubMedQA ($n{=}1{,}000$). Random baseline: 33.3\%. 95\% Wilson score CIs are $\pm$2.0--3.1\,pp for all conditions; see Appendix~\ref{app:trunc} for detailed intervals.}
\label{tab:evidence}
\begin{tabular}{lccl}
\toprule
\textbf{Condition} & \textbf{4B} & \textbf{27B} & \textbf{Description} \\
\midrule
\multicolumn{4}{l}{\textit{Core conditions:}} \\
Question only & 34.5\% & 31.0\% & No context \\
Full context & 45.8\% & 38.2\% & Complete abstract \\
Front-trunc.\ 50\% & 13.8\% & 23.4\% & First half of words \\
Front-trunc.\ 25\% & 13.7\% & 18.6\% & First quarter \\
Background only & 28.9\% & 19.8\% & Intro.\ sentences \\
Results only & 41.9\% & \textbf{40.0\%} & Concl.\ sentences \\
\midrule
\multicolumn{4}{l}{\textit{Expanded truncation strategies (4B):}} \\
Back-trunc.\ 50\% & \textbf{44.5\%} & \textit{--} & Last half of words \\
Middle-trunc.\ 50\% & 33.9\% & \textit{--} & Central portion \\
Sent.-trunc.\ 50\% & 30.3\% & \textit{--} & Sentence boundary \\
Salient top-5 & 40.1\% & \textit{--} & By question overlap \\
Salient top-3 & 36.4\% & \textit{--} & By question overlap \\
\bottomrule
\end{tabular}
\end{table}

\subsection{Robustness Baselines}

Table~\ref{tab:baselines} compares three robustness-oriented methods against zero-shot direct prompting (47.6\%).

\textbf{Cloze scoring} achieves 51.8\% [50.3, 53.3] on 4B and 64.5\% [61.4, 67.1] on 27B, the highest accuracy of any method we tested for both model sizes. The 27B result is striking: cloze scoring recovers 26.3 points over 27B's full-context evidence conditioning (38.2\%) and 17.0 points over 4B's best prompting result (47.6\%). Position bias scores are 0.013 (4B) and 0.054 (27B), an order of magnitude lower than zero-shot direct (0.137).

\textbf{Permutation voting} (4 orderings, $n{=}1{,}000$) achieves 49.0\% [46.0, 52.1] aggregated accuracy, a 4-point improvement over the mean per-permutation accuracy of 45.1\%. The agreement rate across orderings is 70.0\%, with disagreement identifying items where position bias drives predictions.

\begin{table}[t]
\centering
\caption{Robustness baselines on MedMCQA. Cloze scoring uses $n{=}4{,}183$ (4B) or $n{=}1{,}000$ (27B); permutation vote uses $n{=}1{,}000$.}
\label{tab:baselines}
\begin{tabular}{lcccc}
\toprule
\textbf{Method} & \multicolumn{2}{c}{\textbf{MedGemma-4B}} & \multicolumn{2}{c}{\textbf{MedGemma-27B}} \\
\cmidrule(lr){2-3} \cmidrule(lr){4-5}
 & Acc. & Pos.\ Bias & Acc. & Pos.\ Bias \\
\midrule
Zero-shot direct (ref.) & 47.6\% & 0.137 & -- & -- \\
\midrule
Cloze scoring & 51.8\% & 0.013 & \textbf{64.5\%} & 0.054 \\
Permutation vote ($K{=}4$) & 49.0\% & -- & -- & -- \\
\bottomrule
\end{tabular}
\end{table}

The cloze result deserves attention. By bypassing generation entirely and reading the model's internal token preferences, we recover 4.2 points of accuracy that are lost to the generation and parsing pipeline. This suggests that a significant fraction of MedGemma's apparent ``errors'' under standard prompting are not errors of knowledge but of output formatting.

\section{Discussion}

\subsection{Why Chain-of-Thought Hurts}

The 5.7\% accuracy drop from CoT prompting aligns with recent findings that deliberation can reduce performance on certain tasks~\cite{sprague2024mind}. MedGemma was trained extensively on medical text and may have internalized domain reasoning patterns. Forcing explicit step-by-step logic may override these learned patterns with less reliable deliberation.

Case-level analysis reveals the mechanism. CoT prompting changed answers on 1,262 of 4,183 questions, hurting 750 (direct correct, CoT wrong) while helping only 512, a net loss of 238 questions. The predominant failure pattern involves verbose reasoning (90.7\% exceeded 500 characters) where longer chains create opportunities for errors to compound. We also observed self-contradiction in 25.6\% of cases and confident wrong conclusions in 11.1\%.

The characteristic failure mode works like this: the model correctly identifies relevant medical concepts early in its reasoning, considers alternatives, then talks itself into the wrong answer. In one case involving organophosphate poisoning, CoT correctly identified the condition and atropine's role as antidote, then continued deliberating and selected neostigmine, which would worsen the condition.

\subsection{The 59\% Flip Rate Problem}

MedGemma changes its answer 59.1\% of the time when options are shuffled, far exceeding random noise. The maximum flip rate of 72.9\% for rotation perturbations means that for nearly three-quarters of questions, answers depend more on option position than content. This magnitude exceeds typical findings~\cite{zheng2024llm}, suggesting medical-specialist training may not mitigate position bias and could even exacerbate it.

For clinical applications, this fragility is unacceptable. A diagnostic support system that changes recommendations based on option ordering provides no reliable signal to clinicians.

\subsection{Truncation Direction Matters More Than Amount}

Front-truncating context to 50\% yields 13.8\% accuracy while no context achieves 34.5\%, a 20.7-point gap showing that partial context actively misleads. But our expanded analysis reveals this is specifically a \textit{front-truncation} problem: back-truncation at the same 50\% ratio preserves 44.5\% accuracy, just 1.3 points below full context. The model relies on opening context for orientation; once it has read the beginning of an abstract, losing the tail is relatively harmless.

This asymmetry has direct implications for RAG systems in medicine. Prior work shows that models can be led astray by incorrect retrieved evidence~\cite{wu2024clasheval,barnett2024rag}. Our results add specificity: the \textit{position} of missing information matters as much as its quantity. RAG systems that truncate passages to fit context windows should truncate from the end, not the beginning. Better still, salient-sentence extraction (40.1\% with top-5 sentences) provides a principled compression that retains 88\% of full-context accuracy at roughly half the token cost.

Results-only context (41.9\% for 4B, 40.0\% for 27B) nearly matches or exceeds full context, while background-only achieves just 28.9\%/19.8\%. Both models benefit from conclusions rather than methodological background, suggesting RAG systems should prioritize high-information-density content over exhaustive retrieval.

\subsection{Scale and Robustness}

MedGemma-27B underperforms 4B on generative evidence conditioning (38.2\% vs 45.8\% with full context), apparently demonstrating that medical benchmark performance does not scale with model size. But our cloze scoring results invert this conclusion: 27B achieves 64.5\% via log-probabilities, far exceeding 4B's 51.8\%. The 27B model \textit{does} encode substantially more medical knowledge than 4B; its generation pipeline simply fails to express it. This finding reframes the ``scale doesn't help'' narrative. Scale helps knowledge acquisition, but the generation-parsing pipeline becomes the bottleneck at larger sizes.

On evidence conditioning, 27B shows a ``less is more'' pattern: results-only context (40.0\%) exceeds full-context accuracy (38.2\%), suggesting larger models may be more susceptible to distraction from verbose context but respond well to concentrated information.

\subsection{Cloze Scoring as a Diagnostic}

The advantage of cloze scoring over zero-shot direct prompting, 4.2 points for 4B (51.8\% vs.\ 47.6\%), is striking because both methods use the same model and the same questions; the only difference is \textit{how} the answer is extracted. The effect is even more dramatic for 27B: cloze scoring achieves 64.5\%, making 27B the best-performing configuration in our entire study when answers are extracted via log-probabilities rather than parsed from generated text.

Two mechanisms likely contribute to this gap: (1)~the generation pipeline adds noise through token-by-token autoregressive decoding and output formatting constraints; and (2)~the answer parser fails on some responses that contain correct reasoning but not a cleanly extractable letter. The near-zero position bias (0.013 for 4B, 0.054 for 27B vs.\ 0.137 for zero-shot direct) supports this interpretation; the models' internal preferences are far more uniform across positions than their generated text.

This has practical implications. For applications where only the answer matters (not the reasoning), cloze scoring offers a strictly better extraction method. For applications requiring explainability, the cloze-generation gap can serve as a diagnostic of how much accuracy is lost to the generate-parse pipeline.

\subsection{Permutation Voting Partially Mitigates Position Bias}

Permutation voting improves aggregated accuracy by 4 points over the mean single-ordering accuracy (49.0\% vs.\ 45.1\%), confirming that majority-vote aggregation across orderings can partially cancel position-dependent errors. The 70\% agreement rate identifies a natural partition: questions with high agreement are reliably answered regardless of ordering, while low-agreement questions flag items where the model's answer depends on superficial cues. In a clinical deployment, the agreement rate could serve as a calibrated abstention signal.

\subsection{Generalizability and Clinical Safety}

Our experiments focus on the MedGemma family. Whether these ``backfire'' effects generalize to other domain-specific models, such as BioGPT, BioMistral~\cite{labrak2024biomistral}, or clinical adaptations of Llama, remains an open question. Prior work has documented position bias and CoT sensitivity in general-purpose LLMs~\cite{zheng2024llm,sprague2024mind}, so these failure modes are likely not unique to MedGemma. But the magnitudes we observe (e.g., 59.1\% flip rate, 5.7\% CoT degradation) may well differ across architectures and training regimes. Future multi-model evaluations should test whether domain-specific fine-tuning amplifies or mitigates these effects.

Our evaluation also covers only multiple-choice and yes/no/maybe formats. Real clinical workflows involve free-text reasoning, open-ended differential diagnosis, and structured decision support. The mechanisms we identify, including position bias, verbose reasoning errors, and context truncation sensitivity, are likely to persist in those settings, though they may manifest differently.

Prompt-induced prediction changes are not uniformly benign. Among the 750 questions where CoT flipped a correct answer to an incorrect one, a subset involves high-risk clinical scenarios. In the organophosphate poisoning example, CoT selected a contraindicated drug. We observed multiple cases where prompt choice alone led to selection of contraindicated medications (e.g., neostigmine for organophosphate poisoning, allopurinol during acute gout; see Appendix~\ref{app:cot}). While a comprehensive clinical severity taxonomy is beyond our scope, these examples demonstrate that prompt-induced errors can be clinically dangerous---not merely statistically inconvenient---underscoring the need for robustness testing before any safety-critical deployment.

\section{Conclusion}

Our evaluation demonstrates that standard prompt engineering techniques do not reliably transfer to domain-specific medical LLMs. On MedGemma, CoT decreases accuracy by 5.7\% ($p < 0.001$), few-shot examples degrade accuracy by 11.9\% while tripling position bias, option shuffling causes 59.1\% prediction flips, and front-truncated context performs worse than no context at all. Practical mitigations exist: cloze scoring achieves the highest accuracy (51.8\%/64.5\%) with near-zero position bias, back-truncation preserves 97\% of full-context accuracy, and permutation voting recovers 4 points over single-ordering inference.

For deployment, we recommend: (1)~default to zero-shot direct prompting or cloze scoring; (2)~test option order sensitivity and use permutation voting with agreement-based abstention; (3)~truncate retrieved context from the end, not the beginning; and (4)~prefer selective high-density retrieval over exhaustive context. The cloze--generation gap suggests that benchmark accuracies reflect the prompt--generate--parse pipeline rather than underlying model knowledge, making rigorous per-use-case validation essential before deployment.

\subsubsection{Acknowledgements.}
We thank the MedGemma team at Google for releasing open model weights. Experiments were conducted on NVIDIA A100 GPUs.

\bibliographystyle{splncs04}
\bibliography{references}

\appendix

\section{Prompt Templates}\label{app:prompts}

This section provides the exact prompt templates used in our experiments.

\paragraph{Zero-shot Direct.}
\begin{verbatim}
Answer the following medical question by selecting
the correct option.

Question: {question}

A. {option_a}  B. {option_b}
C. {option_c}  D. {option_d}

Answer with just the letter (A, B, C, or D):
\end{verbatim}

\paragraph{Zero-shot Chain-of-Thought.}
\begin{verbatim}
Answer the following medical question.
Think step by step.

Question: {question}

A. {option_a}  B. {option_b}
C. {option_c}  D. {option_d}

Let's think step by step, then provide the
answer as a single letter (A, B, C, or D):
\end{verbatim}

\paragraph{Few-shot Direct.}
Three example questions with correct answers were prepended to each test question. Examples were randomly sampled from different medical subjects to avoid information leakage.

\paragraph{Answer-only.}
\begin{verbatim}
{question}
A. {option_a}  B. {option_b}
C. {option_c}  D. {option_d}
\end{verbatim}

\paragraph{PubMedQA.}
\begin{verbatim}
Based on the following context, answer the
research question.

Context: {context}
Question: {question}

Answer with one word: yes, no, or maybe.
\end{verbatim}

\section{Detailed Chain-of-Thought Analysis}\label{app:cot}

\paragraph{Case-level Breakdown.}
Of 4,183 MedMCQA questions evaluated with both zero-shot direct and zero-shot CoT:

\begin{table}[t]
\centering
\caption{Case-level comparison of direct vs.\ CoT prompting.}
\label{tab:cot_cases}
\begin{tabular}{lrr}
\toprule
\textbf{Outcome} & \textbf{Count} & \textbf{Pct.} \\
\midrule
Both correct & 1,511 & 36.1\% \\
Both wrong & 1,410 & 33.7\% \\
Direct correct, CoT wrong & 750 & 17.9\% \\
Direct wrong, CoT correct & 512 & 12.2\% \\
\midrule
\textbf{Net effect of CoT} & $-$238 & $-$5.7\% \\
\bottomrule
\end{tabular}
\end{table}

\paragraph{Failure Patterns.}
We analyzed 200 cases where CoT hurt performance (Table~\ref{tab:cot_failures}). The predominant pattern is verbose reasoning (90.7\%) where longer chains create opportunities for errors to compound. Categories are not mutually exclusive.

\begin{table}[t]
\centering
\caption{Failure patterns in CoT responses ($n{=}200$ sample).}
\label{tab:cot_failures}
\begin{tabular}{lrl}
\toprule
\textbf{Pattern} & \textbf{\%} & \textbf{Description} \\
\midrule
Verbose reasoning & 90.7 & $>$500 chars \\
Correct concept, wrong applic. & 34.5 & Misapplied knowledge \\
Distractor confusion & 28.0 & Rejected correct ans. \\
Self-contradiction & 25.6 & Conflicting logic \\
Overthinking & 18.5 & Over-complicated \\
Confident wrong concl. & 11.1 & ``Therefore'' + wrong \\
\bottomrule
\end{tabular}
\end{table}

\paragraph{Overthinking Examples.}
Two additional cases illustrate the ``Overthinking'' failure pattern, where CoT introduces unnecessary complexity into straightforward questions.

\smallskip\noindent\textbf{Example A (Pharmacology, correct answer: B).}
\textit{Question:} ``Which drug is used for the treatment of acute gout?''
Options: A)~Allopurinol, B)~Colchicine, C)~Probenecid, D)~Febuxostat.
Direct prompting correctly answers ``B.'' CoT begins: ``Allopurinol is a xanthine oxidase inhibitor used for chronic gout prophylaxis\ldots Colchicine is effective for acute flares\ldots However, NSAIDs are often first-line for acute gout, and colchicine has a narrow therapeutic index\ldots Probenecid is a uricosuric agent\ldots Given the considerations about side effects\ldots'' and selects ``A'' (Allopurinol), a drug contraindicated during acute attacks. The model correctly identified colchicine as the acute treatment but then over-deliberated on side effect considerations irrelevant to the question.

\smallskip\noindent\textbf{Example B (Anatomy, correct answer: C).}
\textit{Question:} ``The longest cranial nerve is:''
Options: A)~Trigeminal, B)~Facial, C)~Vagus, D)~Hypoglossal.
Direct prompting correctly answers ``C.'' CoT responds: ``The trigeminal nerve has three divisions and covers a large facial area\ldots The vagus nerve extends to the abdomen, making it anatomically the longest\ldots However, if we consider the total fiber length including all branches, the trigeminal's three divisions might collectively exceed\ldots'' and selects ``A'' (Trigeminal). The model introduced a specious ``total branch length'' reasoning that turned a factual recall question into an unnecessary deliberation.

\section{Position Bias Analysis}\label{app:bias}

\begin{table}[t]
\centering
\caption{Answer distribution: ground truth vs.\ predictions.}
\label{tab:position_dist}
\begin{tabular}{lcccc}
\toprule
\textbf{Condition} & \textbf{A} & \textbf{B} & \textbf{C} & \textbf{D} \\
\midrule
Ground truth & 32.2\% & 25.1\% & 21.4\% & 21.3\% \\
\midrule
Zero-shot direct & 45.9\% & 22.1\% & 17.8\% & 14.2\% \\
Zero-shot CoT & 52.3\% & 19.4\% & 15.7\% & 12.6\% \\
Few-shot direct & 76.0\% & 12.0\% & 7.0\% & 5.0\% \\
Few-shot CoT & 68.4\% & 15.2\% & 9.8\% & 6.6\% \\
Answer-only & 41.2\% & 24.3\% & 19.1\% & 15.4\% \\
\bottomrule
\end{tabular}
\end{table}

The model shows a consistent preference for option A across all conditions, but this bias is dramatically amplified under few-shot prompting. With few-shot direct, the model predicts A for 76\% of questions despite A being correct only 32.2\% of the time, an overweight of 43.8 percentage points.

\section{Option Order Details}\label{app:order}

\begin{table}[t]
\centering
\caption{Prediction consistency across perturbations.}
\label{tab:consistency}
\begin{tabular}{lr}
\toprule
\textbf{Consistency Level} & \textbf{Pct.} \\
\midrule
All 5 conditions & 23.4\% \\
4 of 5 conditions & 18.7\% \\
3 of 5 conditions & 21.2\% \\
2 of 5 conditions & 19.8\% \\
All different & 16.9\% \\
\bottomrule
\end{tabular}
\end{table}

Only 23.4\% of questions received consistent predictions across all five ordering conditions. For over three-quarters of questions, the model's answer depends on option ordering.

\section{Truncation Strategy Comparison}\label{app:trunc}

\begin{table}[t]
\centering
\caption{Truncation strategies at 50\% reduction (MedGemma-4B, PubMedQA $n{=}1{,}000$).}
\label{tab:trunc_comparison}
\begin{tabular}{lccc}
\toprule
\textbf{Strategy} & \textbf{Acc.} & \textbf{95\% CI} & \textbf{$\Delta$ Full} \\
\midrule
Full context (ref.) & 45.8\% & [42.7, 48.9] & -- \\
\midrule
Back-trunc.\ 50\% & 44.5\% & [41.4, 47.6] & $-$1.3 \\
Salient top-5 & 40.1\% & [37.1, 43.2] & $-$5.7 \\
Middle-trunc.\ 50\% & 33.9\% & [31.0, 36.9] & $-$11.9 \\
Sent.-trunc.\ 50\% & 30.3\% & [27.5, 33.2] & $-$15.5 \\
Front-trunc.\ 50\% & 13.8\% & [11.8, 16.1] & $-$32.0 \\
\midrule
Question only & 34.5\% & [31.6, 37.5] & $-$11.3 \\
\bottomrule
\end{tabular}
\end{table}

The 30.7 percentage point gap between back-truncation (44.5\%) and front-truncation (13.8\%) at the same reduction ratio is the single largest effect in our study.

\section{Robustness Baselines Details}\label{app:baselines}

\paragraph{Cloze Scoring.}
For each MedMCQA question, we extract the log-probabilities of the four option tokens (A, B, C, D) at the final position. The option with the highest log-probability is selected. This requires only a single forward pass with no autoregressive generation.

\begin{table}[t]
\centering
\caption{Cloze scoring detailed results.}
\label{tab:cloze_detail}
\begin{tabular}{lcc}
\toprule
\textbf{Metric} & \textbf{4B ($n{=}4{,}183$)} & \textbf{27B ($n{=}1{,}000$)} \\
\midrule
Accuracy & 51.8\% & 64.5\% \\
95\% CI & [50.3, 53.3] & [61.4, 67.1] \\
Mean logprob margin & 4.07 & 3.14 \\
Position bias & 0.013 & 0.054 \\
\bottomrule
\end{tabular}
\end{table}

\paragraph{Permutation Voting.}
For each question, $K{=}4$ distinct orderings receive independent greedy decoding. Predictions are inverse-mapped to canonical labels; majority vote determines the final answer. The agreement rate serves as a calibration signal.

\begin{table}[t]
\centering
\caption{Permutation voting (MedGemma-4B, $n{=}1{,}000$, $K{=}4$).}
\label{tab:perm_vote_detail}
\begin{tabular}{lc}
\toprule
\textbf{Metric} & \textbf{Value} \\
\midrule
Aggregated accuracy & 49.0\% \\
Per-perm.\ accuracy (mean $\pm$ std) & 45.1\% $\pm$ 1.7\% \\
Mean agreement rate & 70.0\% \\
Agreement rate std & 22.1\% \\
\bottomrule
\end{tabular}
\end{table}

\section{Threats to Validity}\label{app:threats}

\paragraph{Answer Parsing.}
Our evaluation relies on extracting answer letters via regex. Parse error rates are below 2.5\% across all conditions (1.4\% for direct, 2.1\% for CoT). The 0.7-point difference cannot explain the 5.7\% accuracy gap. Moreover, CoT being harder to parse makes our finding more conservative.

\paragraph{Dataset Imbalance.}
The model's overweighting of position A (45.9\% predicted vs.\ 32.2\% true) far exceeds what ground truth would justify. The 59.1\% flip rate from option shuffling directly demonstrates position-based predictions that cannot be explained by dataset imbalance.

\paragraph{Dataset Contamination.}
Both benchmarks are publicly available and may have appeared in pretraining data. However, our analysis focuses on \textit{relative} robustness across conditions. Memorized answers should not change when options are shuffled or prompts are reformatted. The large relative degradations we observe indicate genuine sensitivity independent of potential contamination.

\section{Reproducibility}\label{app:repro}

All randomized experiments use seeds $\{42, 123, 456\}$. MedMCQA outputs are constrained to letters A through D; PubMedQA to yes/no/maybe. Answers are extracted via regex matching the first valid token. Unparseable outputs are scored as incorrect.

All experiment code, prompt templates, configuration files, and analysis scripts are available at: \textit{https://github.com/UNHSAILLab/MedMCQA-Robustness-Study}.

\end{document}